\documentclass[11pt,a4paper]{article}
\usepackage[hyperref]{acl2021}
\usepackage{times}
\usepackage{latexsym}
\usepackage{graphicx}
\usepackage{amsmath}
\usepackage{multirow}
\usepackage{amssymb}
\usepackage{booktabs}
\usepackage{changepage,threeparttable}
\usepackage{booktabs, multirow} % for borders and merged ranges
\usepackage{soul}% for underlines
\usepackage{mathrsfs}
\usepackage{comment}

\newcommand{\rpm}{\raisebox{.2ex}{$\scriptstyle\pm$}}

\usepackage{microtype}

\aclfinalcopy % Uncomment this line for the final submission
\def\aclpaperid{2990} %  Enter the acl Paper ID here

\title{Question Generation for Adaptive Education}

\author{Megha Srivastava \\
  Stanford University \\
  \texttt{megha@cs.stanford.edu} \\\And
  Noah Goodman \\
  Stanford University \\
  \texttt{ngoodman@stanford.edu} \\}

\date{}

\begin{document}
\maketitle
\begin{abstract}
Intelligent and adaptive online education systems aim to make high-quality education available for a diverse range of students. However, existing systems usually depend on a pool of hand-made questions, limiting how fine-grained and open-ended they can be in adapting to individual students. We explore targeted question generation as a controllable sequence generation task. We first show how to fine-tune pre-trained language models for deep knowledge tracing (LM-KT). This model accurately predicts the probability of a student answering a question correctly, and generalizes to questions not seen in training. We then use LM-KT to specify the objective and data for training a model to generate questions  conditioned on the student and target difficulty. Our results show we succeed at generating novel, well-calibrated language translation questions for second language learners from a real online education platform.
\end{abstract}

\section{Introduction}

Online education platforms can increase the accessibility of educational resources around the world.  However, achieving equitable outcomes across diverse learning needs benefits from systems that are adaptive and individualized to each student \citep{doroudi2019fair}. 
Traditionally, adaptive education methods involve planning over a pool of pre-made questions \citep{atkinson1972ed, hunziker2018forget}. These are naturally limited by the diversity and coverage of the pool, as well as the scaling capacity of curriculum planning algorithms. Recent approaches, such as procedural generation for personalized programming games \citep{vargas2017bib}, are limited to well-specified small domains. We address these limitations by leveraging recent success in deep generative models, in particular language models (LMs). 

\begin{figure}[t]
    \centering
    \includegraphics[width=1.0\linewidth]{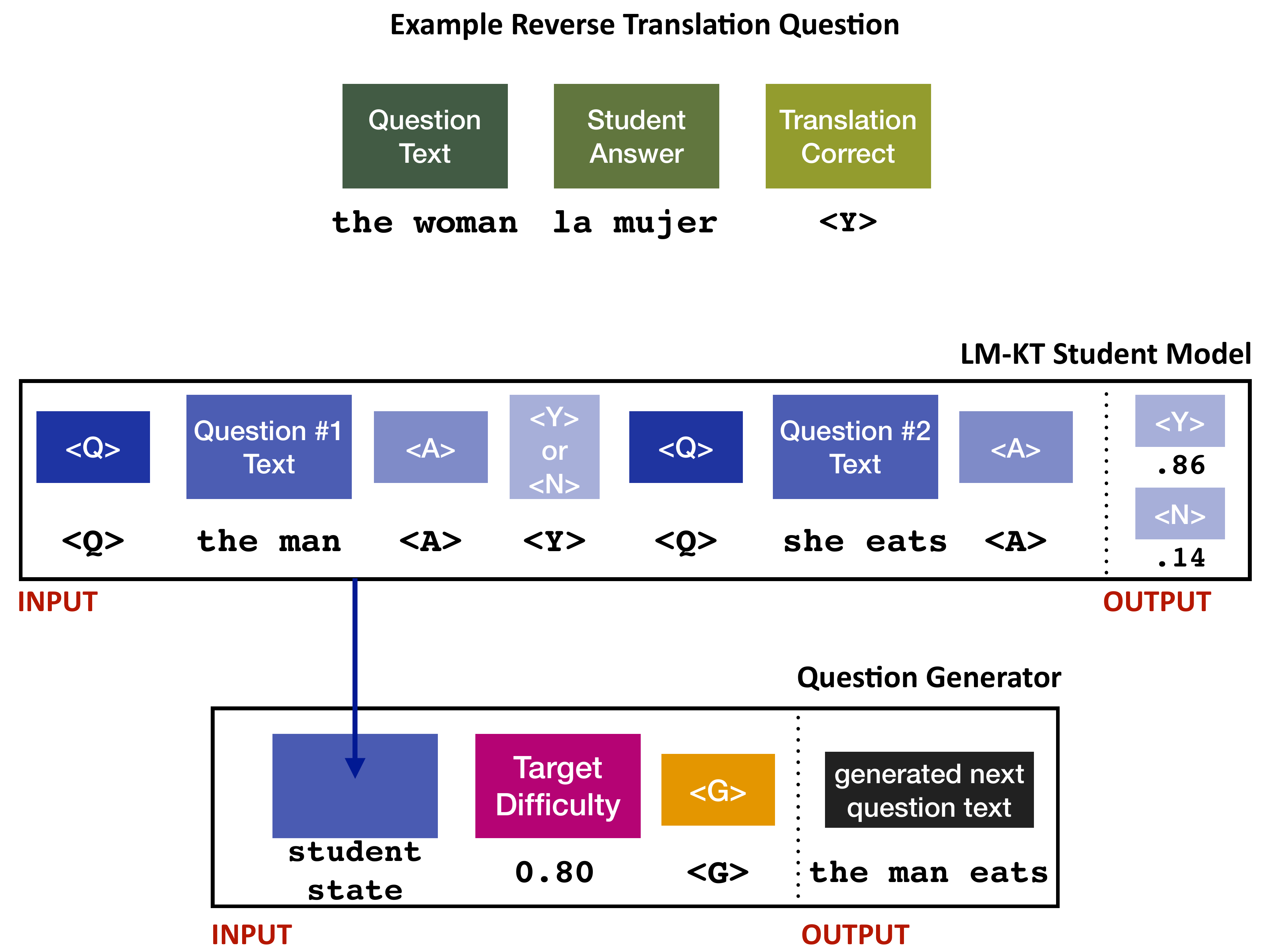}
    \setlength{\belowcaptionskip}{-10pt}
    \caption{Example input and outputs for our LM-based knowledge tracing model (middle) and question generation model (bottom) for an online reverse language translation task (top). A question in this task consists of a target phrase for the student, in this case a Spanish learner, to translate (e.g. ``the woman'').}
    \label{fig:model}
\end{figure}

Many educational activities involve sequential data, such as language translation, reading comprehension, algebra, and deductive logic. Meanwhile, pre-trained LMs can effectively handle sequences from a wide range of modalities \citep{madani2020progen, polu2020generative}. In this work, we focus on natural language sequences, where recent progress in language modeling has shown great success at capturing abstract properties of language \citep{hewitt-manning-2019-structural, liu-etal-2019-linguistic}. 
Specifically, we show how pre-trained LMs can be easily leveraged to adaptively generate questions for a given student and target difficulty in a \textit{reverse translation task}, using difficulty at answering questions as a proxy for more complex future learning objectives. 

We  introduce an LM-based knowledge tracing model (LM-KT) to predict students' difficulty on novel questions (e.g. target phrases to translate). We show that LM-KT is well-calibrated, allowing us to pose the learning problem for the question generator: given a student state, generate a question that will achieve a target difficulty, according to LM-KT. We evaluate both LM-KT and question generation models on real users and responses from  Duolingo\footnote{http://duolingo.com}, a popular online second-language learning platform. 

\section{Background \& Related Works}
There exists a rich body of work on precisely modeling student ``ability" and learning.  For example, Item Response Theory (IRT) seeks to model individual student ability based on their responses to different questions, creating a strong factorization between students and test items \citep{lord1980irt, hambelton2003irt}. Meanwhile, Computer Adaptive Testing (CAT)  techniques are used to determine a fixed student ability as quickly as possible by selecting test items based on information utility \citep{weiss1984kings, thissen2000test, settles-etal-2020-machine}.  However, these methods, which have been used to develop efficient standardized tests, do not necessarily optimize a student's \textit{learning} experience \citep{mu2018adapt}. 
We instead focus on tracking each student's evolving knowledge, choosing questions to target difficulty.

\textbf{Knowledge Tracing} (KT) seeks to model a student's knowledge state from their answer history in order to help individualize exercise sequences \citep{corbett95kt}. This draws inspiration from traditional education curriculum practices, such as distributed spacing of vocabulary \citep{bloom81l2} and mixed review in mathematics \citep{rohrer2009space}. 
To address simplifying assumptions in earlier KT approaches, such as discrete knowledge representations, \citet{piech2015dkt} introduced Deep Knowledge Tracing (DKT), which uses RNNs to enable more complex knowledge representations for students. Recently, SAINT+ \citep{shin2020saint} showed state-of-the-art performance on the popular EdNet KT task using a Transformer model to capture temporal information across activities, motivating our use of Transformer LMs. 

\paragraph{Controllable Text Generation} aims to steer LMs towards desired attributes. Examples include using reinforcement learning to control quality metrics \citep{ranzato2016sequence}, adjusting sampling weights to control for poetry style \citep{ghazvininejad-etal-2017-hafez}, and learning to condition on valence or domain-specific codes \citep{keskar2019ctrl, peng-etal-2018-towards}. To the best of our knowledge, we are the first to use controllable generation in an education context with real student interaction data.

\section{Method}
Given any autoregressive language model (e.g. GPT-2 \citep{radford2019language}, we can fine-tune a \textbf{LM-KT model ($p_{\theta_{KT}}$)} to predict
whether an individual student will correctly answer the next question.
If this model has well-calibrated uncertainty, we can use its predicted probability of a correct answer as a proxy for the difficulty of a question to a student. 
We then train a \textbf{question generation model ($p_{\theta_{QG}}$)} to generate a new question conditioned on a student and desired target difficulty.

\paragraph{Question Representation} Unlike standard DKT, which treats questions as IDs or simple hand-crafted features, we represent questions fully in text (e.g. ``she eats" in Figure \ref{fig:model}). This is a key contribution of our work, required by our eventual goal of \textit{generating} questions in text, and allows the model to leverage similarity across linguistic features. We thus represent a question $q$ as a sequence of words, with prefix and suffix tokens: 
\begin{equation*}
 q_i =\ \  \texttt{<Q>}\ \  w_1^i \ \  w_2^i \ \  w_3^i \ \  ... \ \  w_n^i\ \  \texttt{<A>}
\end{equation*}

\paragraph{Student State}
We represent a student as a temporally-evolving sequence of questions and their responses.
As in much previous KT work, we represent the student response as simply correct/incorrect, with special tokens \texttt{<Y>} and \texttt{<N>}. A student's current state is thus represented as a sequence of all past question and response pairs:
\begin{equation*}
\begin{aligned}
 s_j =\ \  q_1^j\ a_1^j\ q_2^j\ a_2^j \ ...\ q_m^j\ a_m^j\
 \text{, $a_i \in \{$\texttt{<Y>},\texttt{<N>}\}}
 \end{aligned}
\end{equation*}

\paragraph{LM-KT} Given the sequential nature of student learning over time, we can easily frame knowledge tracing as an autoregressive language modeling task. Given a dataset $D$ of students $s_1$, $s_2$, ..., $s_{|D|}$, we employ the standard training objective of finding the parameters $\theta_{KT}$ that minimizes
\begin{equation}
\begin{aligned}
\mathscr{L}_{KT} = - \sum_{i=1}^{|D|}\sum_{t=1}^{|\textbf{x}^{(i)}|}\text{log}p_{\theta_{KT}}(x_t^{(i)}|x_{<t}^{(i)})
 \end{aligned}
\end{equation}
where $\textbf{x}^{(j)}$ $=(x_1^{(j)}, ...., x_{|\textbf{x}|}^{(j)})$ is the entire sequence tokens corresponding to student $s_j$, consisting of all their past questions and answers. Using the softmax output of the LM-KT model ($p_{\theta_{KT}}$), we estimate a student's (inverse) difficulty in answering a specific question  as $d_{qs} = p_{\theta_{KT}}(\texttt{<Y>}|s, q)$. We find that $p_{\theta_{KT}}$ is well-calibrated (Section \ref{sec:sm}), yielding a good proxy for the true question difficulty.

\paragraph{Question Generation} We frame question generation as finetuning a \textit{new} autoregressive LM. Given random samples of students and questions from a held-out set not used to train LM-KT, we can construct a new dataset $D'$ consisting of $s_i \ d_i \texttt{<G>}\ q_i$ sequences, where \texttt{<G>} is a special generation token and $d_i = p_{\theta_{KT}}(\texttt{<Y>}|s_i, q_i)$ is the continuous difficulty value assigned by LM-KT. 
We learn a linear layer to map the continuous input difficulty into a  \textit{difficulty control vector $c_d$} of dimension matching the LM word-embeddings, which we append to the token embeddings.
Unlike LM-KT, we train our question generation model $p_{\theta_{QG}}$ to minimize the loss only on the question text, which only appears \textit{after} the\texttt{<G>} token. If $t_g$ is the token index of \texttt{<G>}, then our modified loss is:
\begin{equation}
\begin{aligned}
\mathscr{L}_{QG} = - \sum_{i=1}^{|D'|}\sum_{t=t_g+1}^{|\textbf{x}^{(i)}|}\text{log}p_{\theta_{QG}}(x_t^{(i)}|x_{<t}^{(i)})
 \end{aligned}
\end{equation}
where sequence $\textbf{x}^{(j)}$ contains the full $s_j \ d_j \texttt{<G>}q_j$ sequence. At test time, we generate tokens $w_1 ... w_n$ conditioned on the $s_j \ d_j \ \texttt{<G>}$ prefix. 
 \label{sec:setup}

\section{Experiments}
Our method generalizes to any education activity that can be represented with text sequences. Due to the availability of real student learning data, we focus on a \textit{reverse language translation} task, where a student translates phrases from their native language (e.g. English, ``she eats") to the second language they are learning (e.g. Spanish, ``ella come'').

\subsection{Experimental Details}

We use the 2018 Duolingo Shared Task on Second Language Acquisition Modeling \citep{slam18} dataset, which contains questions and responses for Duolingo users over the first 30 days of learning a second language. While the original task's goal was to identify \textit{token}-level mistakes, we collapse these errors into binary (correct / incorrect)  per-question labels.
We use the provided train/dev/test splits for users learning Spanish and French. We create separate held-out sets from the test set to evaluate the LM-KT and question generation models. For both models, we finetune separate GPT-2 \citep{radford2019language} models. While we sample from a held-out set of student states and questions to train the question generation model, in principle questions can come from any source text domain. Further experiment details are in the Appendix, and source code can be found at: \url{https://github.com/meghabyte/acl2021-education}.

\subsection{Results: Student Modeling} \label{sec:sm}
We evaluate LM-KT two ways: first, its ability to predict if an individual student will answer a novel question correctly on a held-out test set of real Duolingo student responses. Second, how well-calibrated these predictions are, which is crucial to our later use of LM-KT for question generation.

\begin{table}
\centering
\footnotesize
\begin{tabular}{lllll}
\hline 
\textbf{Model} \textit{(Spanish)} & \textbf{AUC (seen)}& \textbf{AUC (unseen)} \\
\hline
\textbf{LM-KT} & \textbf{0.75 \rpm .0001} & \textbf{0.76 \rpm .001}
\\ 
Standard DKT &  0.72 \rpm .0001  & 0.70 \rpm .001\\ 
Question Only & 0.67 \rpm .0001 & 0.58 \rpm .002\\ \hline 
\textbf{Model} \textit{(French)} & \textbf{AUC (seen) }& \textbf{AUC (unseen)} \\
\hline
\textbf{LM-KT} &  \textbf{0.73 \rpm .0002} & \textbf{0.71 \rpm .002}
\\ 
Standard DKT &  0.70 \rpm .0001  & 0.65 \rpm .002\\ 
Question Only &  0.65 \rpm .0002 &   0.62 \rpm .001\\ \hline
\end{tabular}
\setlength{\belowcaptionskip}{-10pt}
\caption{ LM-KT improves AUC for both questions in the Duolingo test set that were seen during training (for other students) \textit{and} novel questions, over Standard DKT with Question IDs and question-only baselines. Errors are 95\% CIs.}
\label{auc}
\end{table}

Table \ref{auc} compares AUC-ROC on a held-out test set for our LM-KT model with standard DKT, which uses question IDs instead of text, and a baseline that ignores the student state, only using the question text representation. This question only baseline would perform well if the Duolingo dataset largely consisted of universally ``easy'' and ``difficult'' questions, independent of individual student. Our results show that incorporating the student state is crucial for accurately predicting Duolingo user responses, and including question text also leads to a significant improvement. LM-KT outperforms Standard DKT especially on novel questions---a necessary generalization ability for generation.

\begin{figure}[t]
    \centering
    \includegraphics[width=0.45\textwidth]{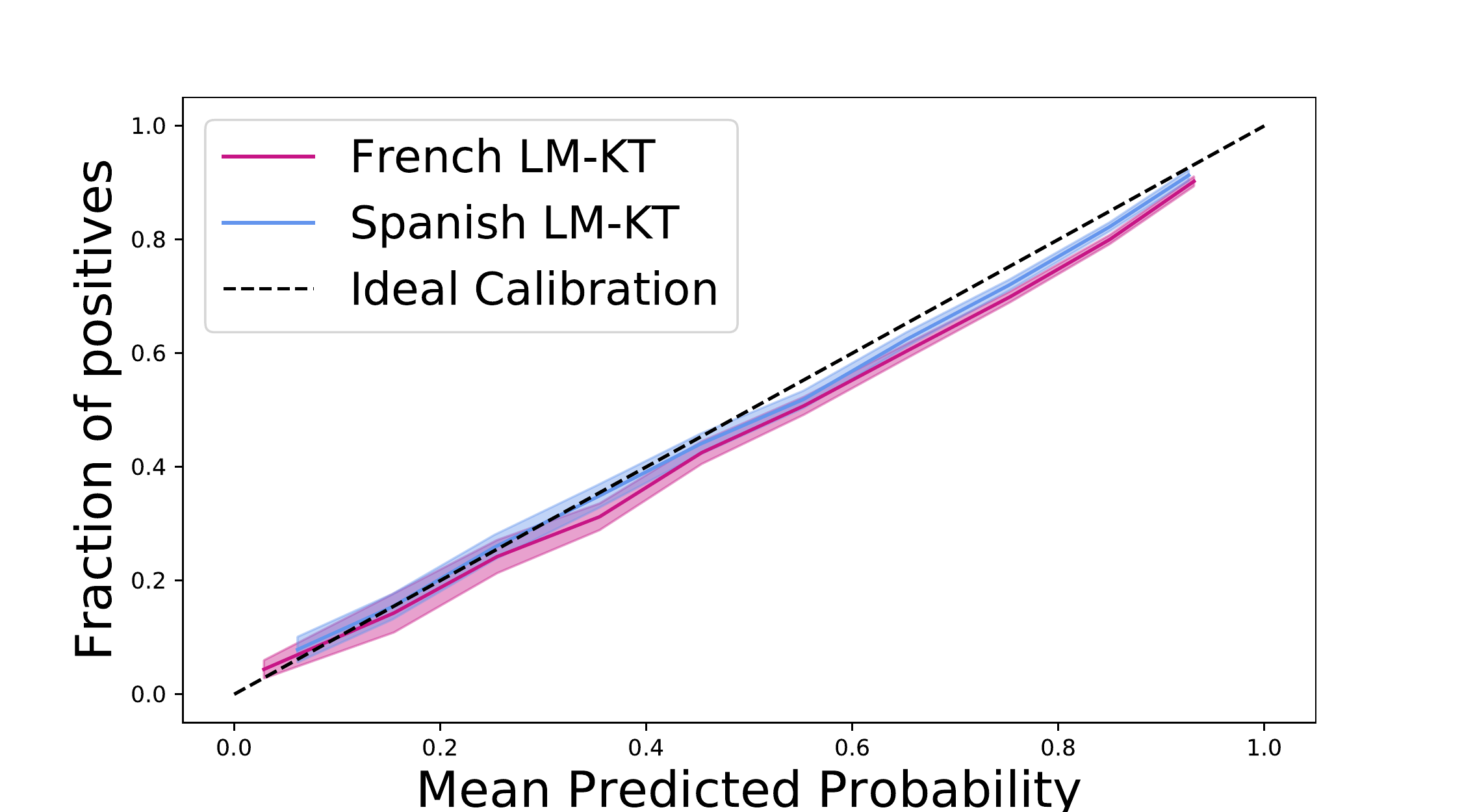}
    \setlength{\belowcaptionskip}{-10pt}
    \caption{Both LM-KT models are well calibrated, but the French model is slightly more overconfident. Filled area shows bootstrap (n=1000) standard deviation.}
    \label{fig:calibrationl}
\end{figure}

Finally, we measure the calibration of our LM-KT models for both Spanish and French (from English) learners, which is the crucial property for our downstream generation task. We bin our test data by predicted question difficulty, and plot the fraction of true correct answers in each bin. Figure \ref{fig:calibrationl} shows that LM-KT is well-calibrated, for both Spanish and French, meaning the predicted difficulty matches the empirically observed proportion of correct answers.

\begin{comment}
\begin{table}
\centering
\footnotesize
\begin{tabular}{lcr}
\hline \textbf{Translation Questions} && \textbf{Difficulty} \\\hline
\textit{Spanish Learning Task} \\ \hline
what? && 0.93 \rpm 0.02   \\
the soap is white && 0.48 \rpm 0.23 \\
we eat sugar in december && 0.14 \rpm 0.16 \\
\hline
\textit{French Learning Task} \\ \hline
no! && 0.959 \rpm 0.03   \\
i eat some baguettes && 0.68 \rpm 0.13 \\
alright, if we can && 0.06 \rpm 0.02 \\
\hline
\end{tabular}
\caption{ Example Duolingo questions and our model's predicted difficulty across five students. Phrases such as ``no!" are considered easy with little variation, while others vary in difficulty based on the student. \ndg{we can leave this table out if we need room.. i don't think it adds much beyond the qualitative description in text.} }
\label{lmktexamples}
\end{table}
\end{comment}

\subsection{Results: Question Generation}
We evaluate four different aspects of our question generation model: (i) successful control for difficulty, (ii) novelty, (iii) fluency, and (iv) latency.     

\paragraph{Difficulty Control}\label{qcontrol} 
To explore whether our question generation model indeed depends on target difficulty and the individual student, we first measure the model's perplexity on a held-out test set of Duolingo questions, compared to permutation baselines.
Table \ref{tab:qppl} (top) shows that perplexity is lower for true student / target difficulty inputs than when either or both of these are permuted.

The target difficulty values in this analysis were defined by the LM-DKT model. We can remove this dependence by using the actual student responses from Duolingo: we set the target difficulty to 1 if the student was correct and 0 otherwise.
Table \ref{tab:qppl} (bottom) shows our model prefers questions paired with these ``true correctness'' targets than paired with random ones.

\begin{figure}[t]
    \centering
    \includegraphics[width=0.45\textwidth]{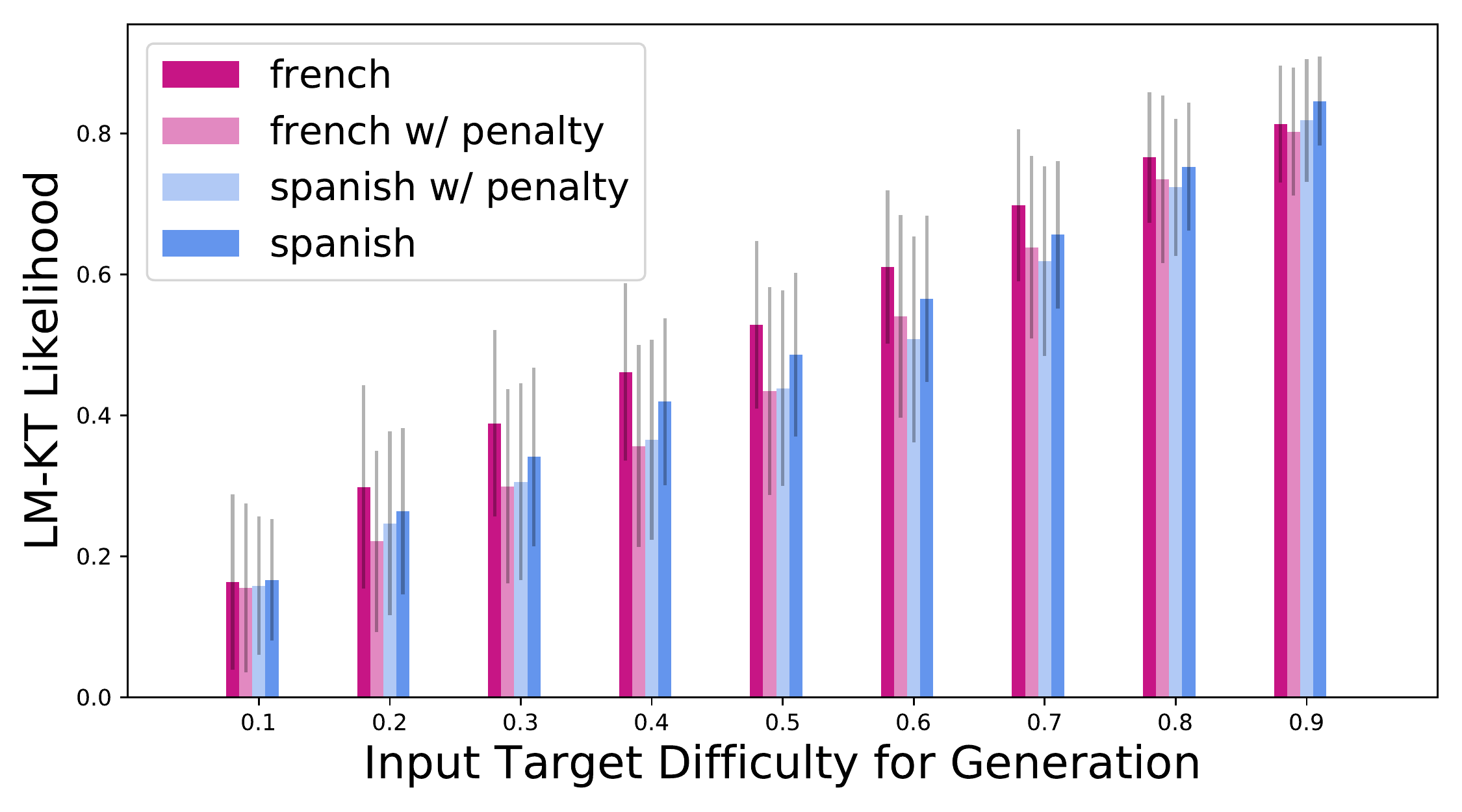}
    \setlength{\belowcaptionskip}{-5pt}
    \caption{For a random selection of 15 students, our question generator successfully controls for difficulty across a range of 9 target values, evaluated by the LM-KT model. Error bars show standard deviation.}
    \label{fig:generationl}
\end{figure}

To evaluate how well our generation model achieves target difficulties, we take 15 unseen students and generate 30 questions for each of 9 input difficulties (0.1-0.9). We then use LM-KT (a well-calibrated proxy for true difficulty) to measure the difficulty of these generated questions for each student. Figure \ref{fig:generationl} shows that we are able to achieve fine-grained control over target difficulty for both Spanish and French students, with an average Root-Mean Squared Error (RMSE) of \textbf{.052} across all students and target difficulties. Adding a sampling penalty \citep{keskar2019ctrl} increases the variance in difficulty (RMSE .062) in exchange for more novel and diverse questions, as discussed next. 

\paragraph{Novelty and Fluency} 
By leveraging a pre-trained language model's ability to manipulate structure, we can generate novel questions not present in the entire Duolingo question set (See Table \ref{tab:examples}). 
Across 4,050 questions generated for Spanish learners,  we found that with a repetition penalty \citep{keskar2019ctrl}, around 43\% of all questions, and 66\% of high difficulty ($d=0.1$) questions, were novel \footnote{The CTRL penalty discounts the scores of previously generated tokens, with the HuggingFace Transformers library \citep{wolf-etal-2020-transformers} implementation including tokens provided as part of the prompt. In our setting, this effectively penalizes for generating questions already seen by the student.}. For French learners, 48\% of all  and 55\% of high difficulty ($d=0.1$) questions were novel. However, around $3 \%$ of generated sentences were judged to be non-fluent \footnote{We use the language-check Python tool to verify grammar https://pypi.org/project/language-check/.}, although most were still able to be translated (e.g.``if i eat some baguettes it breaks."). Without a sampling penalty, the proportion of novel questions drops to about 11 \% of questions for French learners (6 \% for Spanish learners), yet with far fewer non-fluent examples. Further details and examples of novel and non-fluent generated questions for both Spanish and French learners, are in the Appendix.

\begin{table}[h]
\centering
\footnotesize
\begin{tabular}{lll}
\hline
\textbf{Ablation Type} & \textbf{ppl Spanish} & \textbf{ppl French}\\ \hline
\multicolumn{3}{c}{\textit{LM-DKT Likelihood (0 - 1)}} \\
\hline
\textbf{Ground Truth} & \textbf{4.33 \rpm 0.20} & \textbf{3.86 \rpm 0.09} \\
Permute Student & 6.73 \rpm 0.24 & 5.11 \rpm 0.41\\
Permute Difficulty & 12.5 \rpm 1.01 & 7.66 \rpm 0.33 \\ 
Permute Both & 13.1 \rpm 0.43  & 7.87 \rpm 0.26 \\\hline
\multicolumn{3}{c}{\textit{Real Student Answers (0 or 1)}} \\
\hline
\textbf{Ground Truth} & \textbf{17.7 \rpm 1.3} & \textbf{9.49 \rpm .20} \\ 
Permute Student & 19.75 \rpm 0.43 & 10.56 \rpm .60\\
Permute Difficulty & 30.6 \rpm 2.17 & 13.5 \rpm 0.60 \\ 
Permute Both & 31.3 \rpm 1.49 & 13.8 \rpm 0.43 \\ 
\hline
\end{tabular}
\setlength{\belowcaptionskip}{-10pt}
\caption{Perplexity of the question generation model over a held-out evaluation set with ablations.   }\label{tab:qppl}
\end{table}

\begin{table}[h]
\centering
\footnotesize
\begin{tabular}{ll}
\hline
\textbf{Difficulty: 0.1 (very hard)} & \textbf{0.9 (very easy)}\\ 
\hline
 you write letters. & \textcolor{blue}{\textit{spoon or tea}} \\ 
\textcolor{blue}{\textit{i know about that book.}} & socks \\ 
\textcolor{blue}{\textit{she reads your letters.}} & good morning! \\ 
\textcolor{blue}{\textit{those ducks drink water.}}  & a horse  \\
\textcolor{blue}{\textit{he mixes coffee with water.}} & \textcolor{blue}{\textit{oil against salt}}  \\
\hline
\textbf{Difficulty: 0.3 (hard)} & \textbf{0.7 (easy)}\\
\hline
\textcolor{blue}{\textit{you drink juice or water}} & \textcolor{blue}{\textit{what dream?}} \\ 
you drink water & saturday and sunday \\ 
\textcolor{blue}{\textit{accordingly he does it}} & until tomorrow \\ 
\textcolor{blue}{\textit{can we as a band?}}  & \textcolor{blue}{\textit{yes, it is possible!}}   \\
during the night & me too   \\
\hline
\end{tabular}
\setlength{\belowcaptionskip}{-10pt}
\caption{Example questions generated by our model for a Spanish learner. \textcolor{blue}{\textit{Italic questions}} are novel, and do not exist in the Duolingo dataset.}\label{tab:examples}
\end{table}

\begin{figure}[h]
    \centering
    \includegraphics[width=0.45\textwidth]{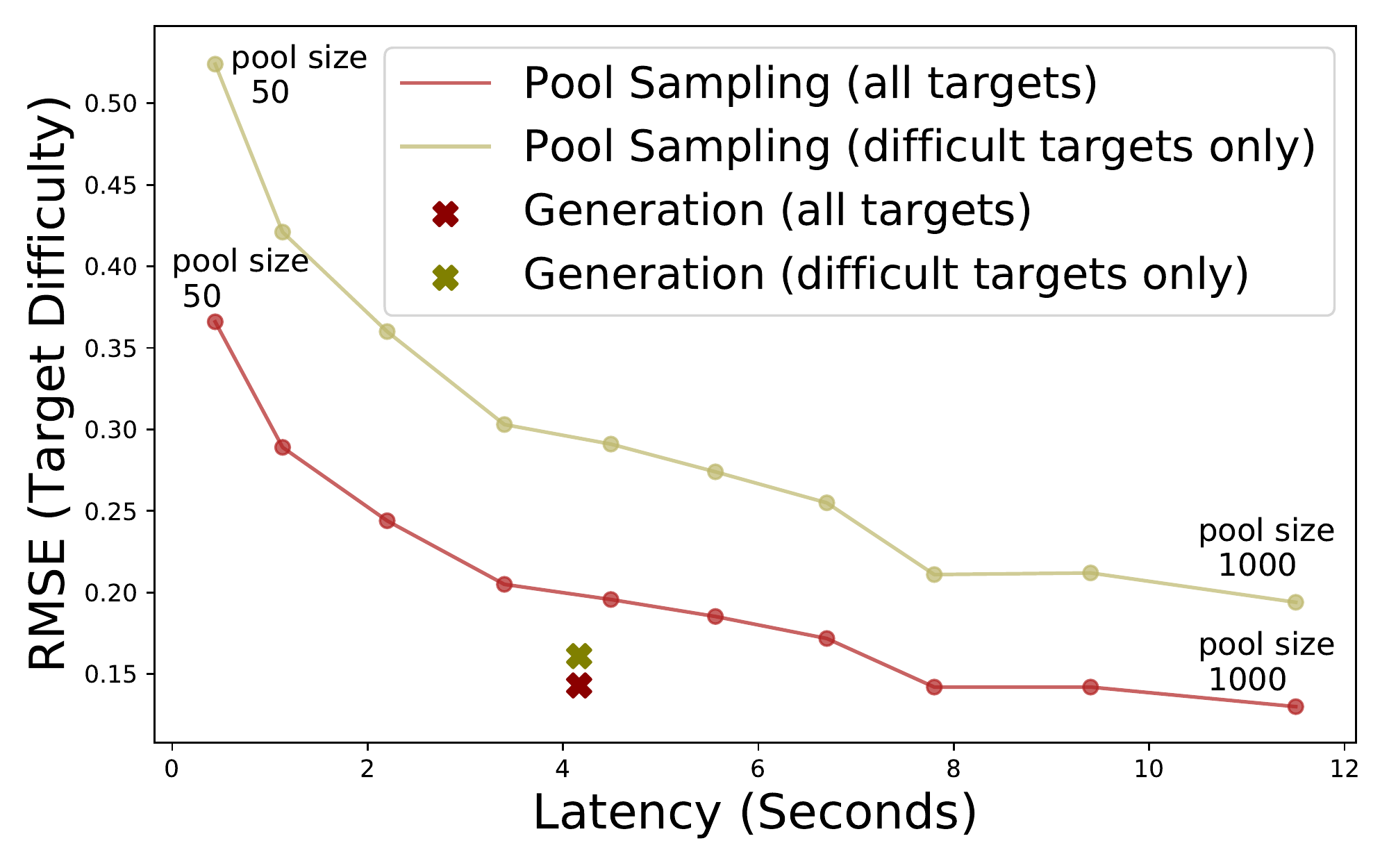}
    \setlength{\belowcaptionskip}{-10pt}
    \caption{Pool selection (for one student) suffers worse question quality vs. latency trade-off than question generation, especially for sampling difficult questions.}
    \label{fig:pooll}
\end{figure}

\paragraph{Latency}
Positive student experience in online education requires low latency.
In about four seconds, our model can generate 30 questions close to a target difficulty.
An alternative to question generation is to rank questions from a preexisting pool, according to a target difficulty objective. We compare the quality (RMSE in achieving target difficulty) of the top 30 questions in a pool against the run-time required to rank all questions in the pool, varying its size (Figure \ref{fig:pooll}). On one NVIDIA Titan XP GPU, we find that, averaged across all target difficulties, 
our question generation model takes half the time to achieve the same quality as pool selection. The gap increases when trying to sample harder questions ( d $<$0.5) -- even a pool size of 1000 does not have sufficient difficult questions, likely due to a skew in the Duolingo question set. Additional controls, such as for style or topic, can easily be combined with our generation method, but would make pool selection exponentially more complex.  

\section{Conclusion}

Our work is a first step toward showing that sequence-based models combined with domain knowledge, such as pre-trained LMs, can be leveraged for adaptive learning tasks. 
We show how to use modern LMs to generate novel reverse-translation questions that achieve a target difficulty, allowing adaptive education methods to expand beyond limited question pools. 

Limitations of our approach include the compute constraints of large LMs and training data availability.
More detailed student data will be crucial to future model development.
For instance, while most publicly available education datasets do not include the full student responses (e.g. full translation response in Duolingo), such information could significantly improve the performance of our LM-KT model. Other future directions include exploring non-language domains, such as math or logic exercises, and  controlling for auxiliary objectives such as question topic.

Finally, designing appropriate user studies to evaluate our method is a complex yet critical next step to determine its suitability in a real-world education setting. Our techniques allows control for individual student difficulty, but it leaves open the question of optimal curriculum design using difficulty-directed question generation.

\section{Broader Impact}
 Online education platforms can increase the accessibility of high quality educational resources for students around the world. Adaptive techniques that allow for more individualized learning strategies can help such technologies be more inclusive for students who make less-common mistakes or have different prior backgrounds \citep{Lee2012KT}. However, our method is subject to biases found in the training data, and careful consideration of using safe and appropriate data is crucial in an education context. Moreover, our specific use of pre-trained LMs relies on the significant progress of NLP tools for English language -- further research and development of these tools for other languages can help ensure our method benefits a larger population of students.  
 
 \section{Acknowledgements} This work was supported in part by the Stanford HAI Hoffman--Yee project ``AI Tutors to Help Prepare Students for the 21st Century Workforce''. MS was additionally supported by the NSF Graduate Research Fellowship Program under Grant No. DGE 1656518.

\bibliographystyle{acl_natbib}
\bibliography{anthology,acl2021}

\newpage
\newpage
\newpage
\newpage
\newpage
\newpage
\newpage
\newpage
\appendix
\newpage
\clearpage 
\onecolumn

\section{APPENDIX}
\subsection{Dataset Details}
The 2018 Duolingo Shared Task on Second Language Acquisition Modeling \citep{slam18} dataset contains questions and responses for Duolingo users over the first 30 days of learning a second language. The dataset contains three different question types: reverse translate (free response translation of a given prompt in the language they are learning), reverse tap (a selection-based equivalent of reverse translate), and listen, where students listen to a vocal utterance. We focus on the reverse translate question type for English-speaking students learning French and Spanish. The dataset size for French learners (1.2k users) is roughly half the size of that for Spanish learners (2.6k users). 

Because the original dataset was intended for per-token error prediction, each question has per-token information that includes whether the student translated the token correctly, as well as Universal Dependencies tags such as part of speech and morphology labels. We use the full question text, rather than individual tokens, for our task, and combine the labels such that if a Duolingo user  incorrectly translated one or more tokens in a question, the entire question is marked  incorrect. We do not use any additional features.  

We use the publicly provided train/dev/test splits from the Shared Task, which are temporally ordered in sequence. We therefore construct student states by tracking user IDs throughout the datasets and appending each new question and response to the current student state. When evaluating our LM-KT model, we use the true responses of preceding questions in the test set to form the student state for a given question. Overall, we find that the dataset is severely imbalanced (as in the original task) - about $30 \%$ of questions are answered incorrectly across students studying both French and Spanish. 

Finally, we create a held-out set of Duolingo questions for both French and Spanish learners to create the training data for our question generation model. From a set of random student states, we select questions from this set and use a trained LM-KT model to assign the difficulty score. In practice, this held-out set can come from any source, not just Duolingo data. 

\subsection{Model Training Details}
To train both our LM-KT knowledge tracing model and our question generation model, we use the pre-trained OpenAI GPT-2 model from the HuggingFace Transformers library \citep{wolf-etal-2020-transformers}. For question generation, we modify the library to add a linear layer and the modified loss function for question generation from Section \ref{sec:setup}. 

We use 1 NVIDIA TitanXP GPU with 12GB of memory available. Because the maximum input sequence length of the GPT-2 model we use is 1024 tokens, we resize all inputs to the last 1024 tokens before training.  We report results for an LM-KT model trained for 13k steps with the default batch size of 2 and learning rate of 5e-5, and a Question Generation model trained for 25k steps with the same batch size and learning rate.  The total compute time to train both models was 2.5 hours for each language learning task.   

\subsection{Question Generation Details}
For both French and Spanish question generation models, we select 15 students unseen during training and generate 30 questions across 9 difficulties from 0.1 to 0.9, using nucleus sampling \citep{holtzman2020curious} ($p=0.99$) with a maximum output length of 20 tokens. We also vary a repetition penalty \citep{keskar2019ctrl} that penalizes for previous tokens (including those in the student state). Lastly, we resize all prompts (student state and target difficulty) to fit into the GPT-2 Model by taking the most recent 1024 tokens, as in training. This is a limitation of our work, as the full student history is not able to be considered for students who have answered a large set of questions. 

\clearpage 
\subsection{Additional Question Generation Outputs}
Our question generation model demonstrates the ability to generate novel questions that do not exist in the entire Duolingo question dataset, especially when a sampling penalty is applied to encourage more diverse outputs. However, this comes at a cost to fluency. Below we include a set of outputs generated by our model for 1 Spanish  student and 1 French student from the Duolingo dataset, with a target difficulty of $d=0.1$, and both with and without a repetition penalty. We observe that while applying a penalty results in a far more novel questions generated, several of these are also non-fluent, using a combination of manual judgement and the Python language-check package (https://pypi.org/project/language-check/).

\begin{table}[!htp]\centering
\caption{Random selection of generated questions for one Spanish learner with for a a target difficulty of $d=0.1$. \textit{Italic questions} are novel, \textbf{bold questions} are judged to be non-fluent.}\label{tab:spanishappx}
\footnotesize
\begin{tabular}{lrr}\toprule
\textbf{Spanish (w/ Penalty)} &\textbf{Spanish (No Penalty)} \\\cmidrule{1-2}
\textit{accordingly he does it.} & \textit{he mixes coffee with milk.} \\\cmidrule{1-2}
\textit{clean your room or close!} & \textit{the cuts are not big.} \\\cmidrule{1-2}
\textit{clean your room!} & \textit{the gallery is enormous.} \\\cmidrule{1-2}
\textit{he mixes coffee with water.} & \textit{the horses are not natural.} \\\cmidrule{1-2}
\textit{how many elephants eat cheese or fish?} & \textit{the men drink a beer.} \\\cmidrule{1-2}
\textit{i know about that book.} &\textit{\textbf{ they probably do not think me.}} \\\cmidrule{1-2}
\textit{ october finds him maximum distance from here today!} & \textit{we can desk a book.} \\\cmidrule{1-2}
\textit{please clean your room!} & from september to december \\\cmidrule{1-2}
\textit{please open your bottle or newspaper?} & according to you, it is yellow. \\\cmidrule{1-2}
\textit{she blames us!} & clean the mirror. \\\cmidrule{1-2}
\textit{she reads us lunchtime newspapers.} & i do not know it. \\\cmidrule{1-2}
\textit{she reads your letters.} & i read the newspaper. \\\cmidrule{1-2}
\textit{those ducks drink water.} & i want a sandwich without cheese. \\\cmidrule{1-2}
\textit{we can abandon him. }& june starts tomorrow. \\\cmidrule{1-2}
\textit{\textbf{ what book have they Chosen me so far?}} & she reads the calendar. \\\cmidrule{1-2}
\textit{you can control her water.} & the plates are not big. \\\cmidrule{1-2}
\textit{you can establish two properties. }& we are following the clue. \\\cmidrule{1-2}
\textit{\textbf{ your house is very put- pretty!}} & we drink quickly. \\\cmidrule{1-2}
previously on television & we eat strawberries. \\\cmidrule{1-2}
you can create the menu. & you can control the water. \\\cmidrule{1-2}
you write letters. & you can create the menu. \\\cmidrule{1-2}
your hat is gray & you can establish a restaurant. \\\midrule
\bottomrule
\end{tabular}
\end{table}

\begin{table}[t]\centering
\caption{Random selection of generated questions for one French learner with for a a target difficulty of $d=0.1$. \textit{Italic questions} are novel, \textbf{bold questions} are judged to be non-fluent.}\label{tab:frenchappdz}
\footnotesize
\begin{tabular}{lrr}\toprule
\textbf{French (w/ Penalty)} &\textbf{French (No Penalty)} \\\cmidrule{1-2}
\textit{do these children have beans?} & \textit{do you have three daughters?} \\\cmidrule{1-2}
\textit{do they come here often? }& \textit{do you like this?} \\\cmidrule{1-2}
\textit{do we come here often or frequently?} & \textit{do you speak french?} \\\cmidrule{1-2}
\textit{do we have chocolate or water?} &\textit{ where do the children read?} \\\cmidrule{1-2}
\textit{do we have coffee here or elsewhere?} & do you come here often? \\\cmidrule{1-2}
\textit{\textbf{ do we have coffee together or onsocks}} & do you want to dance with me? \\\cmidrule{1-2}
\textit{\textbf{ do we like to walk distance from one-to two?}} & some apples, which ones? \\\cmidrule{1-2}
\textit{do we like to walk together or apart?} &corridor or window? \\\cmidrule{1-2}
\textit{do we speak soon or after tomorrow?} &neither do we! \\\cmidrule{1-2}
\textit{is he chinese or Russian?} & you are important. \\\cmidrule{1-2}
\textit{is he chinese or french?} & you are important. \\\cmidrule{1-2}
\textbf{\textit{is he sleeping or going out time?}} & are we going to your place or mine? \\\cmidrule{1-2}
\textit{\textbf{ map ofis suggests an area.} }& corridor or window? \\\cmidrule{1-2}
\textit{\textbf{ otherwise if i want to eat vegetables or fish they regionally cheese, it's meat.}} & do you have a boyfriend? \\\cmidrule{1-2}
\textit{\textbf{ some apples of your apple.}} & do you like to walk? \\\cmidrule{1-2}
\textit{where do we live today? }& neither do we! \\\cmidrule{1-2}
\textbf{\textit{where does he go after that jacket?}} & otherwise, i want a child! \\\cmidrule{1-2}
\textit{where does she go? }& the men are calm and rich. \\\cmidrule{1-2}
\textit{which ones do not fall victim to be sold? }& the parties are in august. \\\cmidrule{1-2}
beans and bread & we are reading your letters. \\\cmidrule{1-2}
corridor or window? & where do we live? \\\cmidrule{1-2}
neither do we! &you eat pork and bread \\\midrule
\bottomrule
\end{tabular}
\end{table}

\end{document}